%% file: paper.tex
\documentclass{article}
\usepackage{spconf,amsmath,graphicx}

\usepackage{epsfig}
\usepackage{xspace}
\usepackage{srcltx}
\usepackage{caption}
\usepackage{subcaption}
\usepackage[export]{adjustbox}
\usepackage{multirow,makecell,hhline}
\usepackage{booktabs}
\usepackage{textcomp}
\usepackage{cite}

\usepackage{url}
\usepackage{hyperref}
\hypersetup{colorlinks=True, urlcolor=black}

\usepackage{color}
\usepackage[dvipsnames]{xcolor}

\newcommand{\CMT}[1]{{}}

\newcommand{\tbh}[1]{\textbf{#1}}

\title{SCALING END-TO-END MODELS FOR LARGE-SCALE MULTILINGUAL ASR}
\name{Bo Li, Ruoming Pang, Tara N. Sainath, Anmol Gulati, Yu Zhang,}
\nameplus{James  Qin, Parisa Haghani, W. Ronny Huang, Min Ma, Junwen Bai\sthanks{Work done when the last author was an intern at Google.}}
\address{Google, USA\\
\texttt{\ninept\{boboli,rpang,tsainath\}@google.com}}
\begin{document}
\ninexpt
\maketitle
\input{0_abstract}

\input{1_intro}

\input{2_system}

\input{3_exps}
\input{4_results}

\input{5_concl}

\section{Acknowledgements}
We would like thank Brian Farris, Chung-Cheng Chiu, Jiahui Yu, Wei Han, Sergey Kishchenko, Ron Weiss and Zhehuai Chen for helpful discussions.

\bibliographystyle{IEEEbib}
\bibliography{mybib}

\end{document}

%% file: 0_abstract.tex
\begin{abstract}
Building ASR models across many languages is a challenging multi-task learning problem due to large variations and heavily unbalanced data. Existing work has shown positive transfer from high resource to low resource languages. However, degradations on high resource languages are commonly observed due to interference from the heterogeneous multilingual data and reduction in per-language capacity. We conduct a capacity study on a 15-language task, with the amount of data per language varying from 7.6K to 53.5K hours. We adopt GShard \cite{lepikhin2020gshard} to efficiently scale up to 10B parameters. Empirically, we find that (1) scaling the number of model parameters is an effective way to solve the capacity bottleneck - our 500M-param model already outperforms monolingual baselines and scaling it to 1B and 10B brought further quality gains; (2) larger models are not only more data efficient, but also more efficient in terms of training cost as measured in TPU days - the 1B-param model reaches the same accuracy at 34\% of training time as the 500M-param model; (3) given a fixed capacity budget, adding depth works better than width and large encoders do better than large decoders; (4) with continuous training, they can be adapted to new languages and domains.
\end{abstract}

%% file: 1_intro.tex
\section{Introduction}
\label{sec:intro}

Large data sets and giant models, together with advances in deep learning algorithms and hardware, enable researchers to push the limit of many machine learning tasks \cite{lepikhin2020gshard,devlin2018bert,brown2020language} including speech \cite{zhang2020pushing}. It brings renewed interests in building universal automatic speech recognition (ASR) systems that can recognize speech from any language. Traditionally, a predefined language agnostic representation is required for building multilingual models such as a global phoneme set like International Phonetic Alphabet (IPA) \cite{international1999handbook}, Speech Assessment Methods Phonetic Alphabet (SAMPA) \cite{wells1995computer} and Worldbet \cite{hieronymus1993ascii} or a universal speech representation like  articulations \cite{frankel2001asr}, which all require expert knowledge. In the past decade, deep neural networks have been widely adopted in the speech community \cite{hinton2012deep}, especially the recently developed end-to-end (E2E) models that merges the modeling of acoustics, lexicon and language into a single neural network \cite{li2020comparison,Ryan19,KimHoriWatanabe17,zhang2020transformer,saon2021advancing}. 
It largely simplifies the building of multilingual models by learning shared representation directly from data. This also speeds up the sharing of techniques between ASR and other machine learning fields such as neural machine translation \cite{arivazhagan2019massively}. Experiments on less than 10 languages have shown promising capabilities of such E2E models in modeling dialects of a particular language \cite{li2018multi}, languages from the same family \cite{kannan2019large} and languages from different families \cite{li2019bytes,pratap2020massively,hou2020large,adams2019massively}. Recently, massively multilingual experiments using more than 50 languages \cite{adams2019massively,pratap2020massively,hou2020large} also show comparable or better performance compared to monolingual systems. 

A major focus of multilingual ASR has been improving performance on low resource languages, which benefit from the data pooling of similar languages, the cross language joint optimization and the consequence positive transfer from higher resource languages \cite{adams2019massively, pratap2020massively, hou2020large, zhou2018multilingual, chuangsuwanich2016multilingual}. However, high resource languages where sufficient monolingual data exists, suffer from interference and constrained capacity \cite{pratap2020massively,wang2020balancing}, and often see a degradation in performance. Improving performance across the board on both high and low resource languages is an under-studied and challenging task. 

From a machine learning perspective, a statistical model's generalization capability builds on inductive bias of the learning algorithms \cite{mitchell1980need}. The underlying inductive bias for multilingual systems is that the learning signal from one language should benefit the quality of other languages \cite{caruana1997multitask}. Under this assumption, the model will generalize better with an increasing number of languages due to the additional information brought by the new languages. This positive transfer is best observed for low resource languages. However, as we increase the number of languages, the modeling task become more challenging due to the large language variations and heavy data imbalance \cite{wang2020balancing}. With a fixed model capacity which is loosely measured in terms of the number of free parameters in neural networks, the positive/negative transfer boundary becomes salient, and high resource languages start to regress due to task interference and reduction in per-task capacity. In \cite{wang2020gradient}, a simple and scalable optimization procedure, namely Gradient Vaccine, is developed to address the gradient interference problem from different tasks. Distilling knowledge from single task models to the multi-task model \cite{li2020knowledge} has also been found to address the interfere problem.

In this paper, we investigate the high resource language performance degradation problem of multilingual models from a capacity perspective \cite{arivazhagan2019massively,lepikhin2020gshard}. Prior work explored as many as 50 \cite{pratap2020massively} to 100 \cite{adams2019massively} languages. However, the scale of the dataset is very limited. The largest language used in \cite{pratap2020massively} has just over 1K hours of speech. In our study, the amount of data per language ranges from 7.6K to 53.5K hours, which leads to high quality monolingual baselines, posing a challenge to build a single multilingual model that can outperform them. We present a capacity solution with thorough empirical studies demonstrating how it is devised. Unlike \cite{pratap2020massively}, no details were provided on how their best 1B model was developed. We adopt the GShard \cite{lepikhin2020gshard} technique to efficiently scale our model up to 10B which shows further word error rate (WER) reductions though relatively small. 
With the increase of model capacity, we manage to recover the performance of monolingual models on all the high resource languages. We ablate the various factors in increasing the model capacity and find that depth generally does better than width and encoder capacity correlates well with recognition performance. We observe that with a fixed capacity, how to feed the language information becomes less important. Moreover, large models are more sample \cite{kaplan2020scaling} and cost efficient, requiring fewer training iterations and less TPU time to reach similar performance. Moreover, we also conducted pilot studies to extend an existing multilingual model to new languages and domains by continuous training, where we found positive knowledge transfers across languages and domains.

%% file: 2_system.tex
\section{Multilingual E2E Models}
\label{sec:system}

\subsection{Model Architecture}
\label{subsec:model}

Our multilingual ASR system is an attention-based encoder-decoder model. For encoder, we use full-context Conformer layers \cite{gulati2020conformer}, consisting of an input projection layer, a relative position embedding layer followed by a stack of Conformer layers  which are organized into three blocks. The first Conformer block consists of 4 Conformer layers followed by a time stacking layer that concatenates the current output with one frame on its left. This doubles the output dim while achieving a 2X time reduction. The second block consists of a single Conformer layer and a projection layer which halves the feature dim and brings it back to the same dimension as the other layers. The remaining Conformer layers make up the third block. Similar to \cite{bo21better}, we use the existing convolution module to provide relative positional information and group normalization to address variations across languages in each Conformer layer.  

We experiment with two types of decoders. Firstly, a unidirectional Long Short-Term Memory (LSTM) \cite{Hasim14} based decoder is used together with a 4-head additive content-based attention to form a Listen, Attend and Spell (LAS) model \cite{Chan15}. Secondly, we also adopt a Transformer decoder with masked self attention and cross attention to the encoder outputs \cite{Vaswani17,zhang2020transformer}. 

The output vocabulary of our multilingual ASR is a unified grapheme set with 3328 tokens; among those, 3,315 tokens appear at least 1000 times in the training data and the remaining are special tokens like ``\textless s\textgreater'', ``\textless /s\textgreater'' and padded placeholders. The majority of the graphemes (3,055) come from Chinese; even with that, Chinese is the only language that has OOV grapheme problem due to the  selection threshold and the coverage of training data. We feed language information via a one-hot embedding vector into the encoder as either an additional input \cite{li2018multi} or the switch for language adapters \cite{li2020knowledge,kannan2019large}. We simply pool all the data together and sample each batch according to the natural distribution.  The whole encoder-decoder network is jointly optimized to minimize the cross-entropy loss between the output of the network and the ground truth transcripts.

\subsection{Scaling Up Model Capacity}

There are multiple ways to scale up an encoder-decoded based multilingual model. In this work, we empirically study the effect of the following factors: {\bf(1)} width {\it vs.} depth; {\bf(2)} encoder {\it vs.} decoder; {\bf(3)} language dependent capacity {\it vs.} language independent capacity; {\bf(4)} architecture {\it vs.} capacity. Strictly speaking, model capacity is not equivalent to the number of parameters {\it i.e.} the model size. For models with language dependent components built in such as adapter models \cite{li2020knowledge}, the inference capacity is smaller than training as during inference only the shared parameters and those corresponding to a specific language are active. To simplify the discussion, we  look at the training model capacity and use model size and capacity interchangeably. 

Scaling up models comes with various practical challenges: the model parallelism support, the computation cost, and the infrastructure support, etc. Recently, the GShard annotation API has been developed for parallel execution \cite{lepikhin2020gshard} and is released in Lingvo \cite{shen2019lingvo} which makes building giant models simpler. Additionally, a new compiler infrastructure, namely the Single Program Multiple Data (SPMD) partitioner , is developed, which makes the compilation time near constant regardless of the number of partitions\cite{lepikhin2020gshard}. This allows us to more efficiently scale up to thousands of partitions. We hence adopt these advances to scale up our multilingual models to 1B parameters and beyond.

%% file: 3_exps.tex
\section{Experimental Details}
\label{sec:exp}

\subsection{Data}

\begin{table}[t]
\caption{Per language training data statistics. Utterance counts are in millions (M) and duration is in thousand (K) hours.}
\hspace{-0.1in}
\begin{tabular}{lcrr}
\toprule
\tbh{Language} & \tbh{Family} & \tbh{Counts}(M) & \tbh{Hours}(K)\\
\midrule
\midrule
English (US) & \multirow{2}{*}{Germanic} & 34.6 & 53.5\\
English (IN) & ~ & 17.9 & 27.1 \\
\midrule
Spanish (US) & \multirow{3}{*}{Italic} & 31.3 & 47.6 \\
Portuguese (BR) & ~ & 17.9 & 32.9 \\
Spanish (ES) & ~ & 16.1 & 23.5 \\
\midrule
Arabic (GULF) & \multirow{2}{*}{Afro-Asiatic} & 7.7 & 11.9 \\
Arabic (EG) & ~ & 7.6 & 11.9 \\
\midrule
Hindi (IN) & \multirow{3}{*}{Indo-Iranian} & 19.8 & 32.3 \\
Marathi (IN) & ~ & 11.4 & 16.7 \\
Bengali (BD) & ~ & 8.6 & 16.5 \\
\midrule
Chinese (TW) & Sino-Tibetan & 17.2 & 22.8 \\
\midrule 
Russian (RU) & Balto-Slavic & 14.8 & 22.8 \\
\midrule
Turkish (TR) & Turkic & 15.5 & 22.1 \\
\midrule
Hungarian (HU) & Uralic & 6.5 & 9.9 \\
\midrule
Malay (MY) & Austronesian & 4.6 & 7.6 \\
\midrule 
\midrule
\multicolumn{2}{l}{\tbh{Total}} & 231.6 & 359.2 \\
\bottomrule
\end{tabular}
\label{tbl:data}
\vspace{-0.2in}
\end{table}

Experiments are conducted on a dataset of 15 languages from 9 distinct language families. Though Germanic, Italic, Indo-Iranian, and Balto-Slavic are members of the Indo-European language family, we treat them as separate language families in this study. There are totally 231.6M utterances which correspond to 359.2K hours of speech data from Google's Voice Search traffic. This is more than 20 times of the data used in \cite{pratap2020massively}. To the best of our knowledge, this is the first work looking at building multilingual ASRs at such a large scale. 
The data is annonymized and human transcribed. Per language data statistics are listed in Table \ref{tbl:data}. The number of utterances for each language ranges from 4.6M to 34.6M, roughly corresponding to 7.6K to 53.5K hours of speech data. The unbalanced data distribution poses a modeling challenge. Unlike other existing multilingual work, we focus on investigating the interference problem between high resource languages. The smallest language in our setup has around 7.6K hours of training data, which is about 7 times of the highest resource language used in \cite{pratap2020massively}. This large scale dataset again bring in training efficiency challenges. The test set for each language contains around 3$\sim$19K utterances sampled from Google's Voice Search traffic with no overlapping from the training set. Similarly, they are anonoymized and hand-transcribed for evaluation purpose.

\subsection{Training}

We use 80D log Mel features that are computed using 32ms windows with a 10ms hop. Features from 3 contiguous frames are stacked and subsampled to form a 240D input representation with 30ms frame rate. A 16D one-hot language vector is fed into the encoder as an additional input. SpecAugment \cite{Park_2019} is used to improve models' robustness against noise. Specifically, two frequency masks with a maximum length of 27 and two time masks with a maximum length of 50 are used. 

All the models are trained in Tensorflow \cite{AbadiAgarwalBarhamEtAl15} using the Lingvo \cite{shen2019lingvo} toolkit on Google's Tensor Processing Units (TPU) V3 with a global batch size of 4,096 utterances. Models are trained with 512 TPU cores except for 10B models which use 1024 TPU cores. This is mainly due to the 16G per core high bandwidth memory (HBM) limit. Models are optimized using synchronized stochastic gradient descent. For LSTM-decoder models, we use the Adam optimizer \cite{KingmaBa15} with parameters $\beta_1$=0.9 and $\beta_2$=0.999; for Transformer-decoder models, Adafactor \cite{shazeer2018adafactor} with parameters $\beta_1$=0.9 and $\beta_2$=0.99 is used. A transformer learning rate schedule \cite{Vaswani17} with peak learning rate 3e-4 and 10K warm-up steps is used.

%% file: 4_results.tex
\section{Results and Discussions}
\label{sec:results}

\begin{figure*}[!t]
\hspace{-0.05in}
    \includegraphics[scale=0.39]{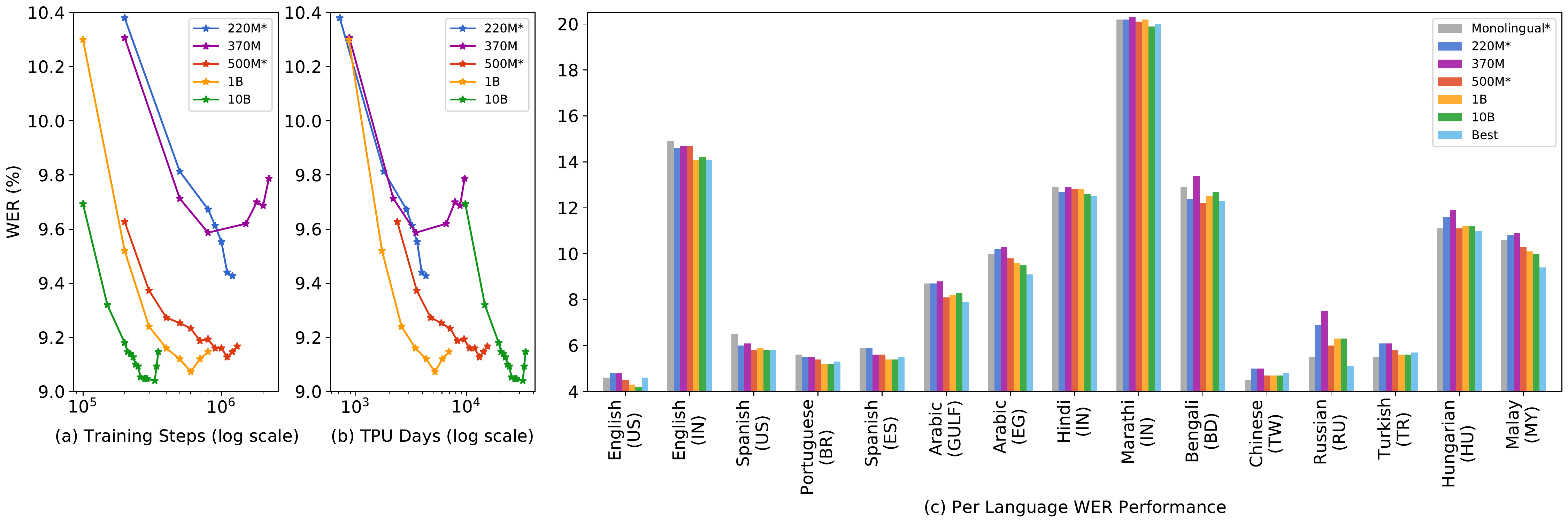}
    \caption{WER performance (\%) vs. {\bf(a)} training steps, {\bf(b)} TPU days and {\bf(c)} language. Systems with * use LSTM decoders.
    }
    \label{fig:final}
\end{figure*}

In this section we present our study of building high quality multilingual models on large scale dataset. For simplicity, we use the average WER for comparisons and only report the per language performance at the end.

\subsection{Monolingual Baselines}

Conformer has been shown to perform the best on many English tasks \cite{gulati2020conformer, bo21better}. We hence adopt it for our monolingual baselines, specifically we use the same Conformer encoder and LSTM decoder architecture as \cite{bo21better} but in a full context setup. The encoder consists of 17 layers of Conformer blocks with 12 layers in the last block ({\it c.f.} Section \ref{subsec:model}). Each Conformer layer has a model dimension of 512. 8-head attention is used in the self-attention layer and the Convolution kernel size used is 15. The decoder is an LSTM based LAS decoder, consisting of 2 layers of 640D LSTM with 2048 hidden units. 4-head content-based additive attention is used in the LAS attention module. Each monolingual model has around 140M parameters and is trained to predict language dependent graphemes only. The average WER is 9.29\% and the per language breakdown is depicted in Figure \ref{fig:final}(c). Across languages, WER ranges from English (US)'s 4.6\% to Marathi (IN)'s 20.2\%. Languages with more data tend to have lower WERs.

\subsection{Multilingual Encoder Architecture}

To justify the effectiveness of Conformer encoders for multilingual modeling, we compare three encoders with the same LSTM-based decoder: {\bf (1) LSTM encoder} with 8 layers, 2048D hidden units and 640D output units; {\bf (2) ContextNet encoder} with 24 layers, 640D hidden units and channel scale of 2 \cite{han2020contextnet}; and {\bf (3) Conformer encoder} with 17 layers, 512D hidden dimension the same as the monolingual baselines.
Language adapters are inserted between each encoder layers. The specific configurations of these three encoders are chosen such that the total number of model parameters are roughly the same, which is around 220M. The increase of model size compared to monolingual models comes from the additional language adapters and the increase of output vocabulary size. 
The average WER of these three models are 11.86\%, 10.77\% and 9.43\% respectively. This clearly demonstrates the effectiveness of Conformer for multilingual ASR. Comparing to monolingual models, even though this baseline multilingual Conformer still lags behind in quality, it does reasonably well in recognizing all the 15 languages. It converges in around 1.2M training steps which roughly corresponds to 21 epochs, while the monolingual models normally train up to 50 epochs.

To understand the effect of the language adapters, we conduct the following ablation studies. For quick experimentation, we compare models at 200K steps (roughly 3.5 epochs), which we find sufficient for model selection. 

{\bf The necessity of language dependent parameters}. The use of language adapters brings in both language dependent model parameters and a small increase in model size. To understand which helps more, we train a single adapter model that forces all the languages to share the same adapter transformation. In this way we can isolate the model size increase from the adapter model. 
This model achieves 10.86\% average WER {\it vs.} the baseline language dependent adapter model's 10.38\% @200K steps. This suggests feeding in language information and learning language dependent parameters are important. 

{\bf The necessity of language dependent transforms}. A simpler way of incorporating language information is to append the one-hot language vector to the input features. It effectively adds language dependent biases; while adapters bring in additional language dependent transformations.
At 200K steps, the bias-only model achieves an average WER of 10.93\% which is worse than the adapter model. However, one thing to note is it has less number of parameters (146M {\it vs.} 220M). We further increase the bias-only model to the same 220M, which yields an average WER of 10.37\% @200K steps similar to the adapter model. Though they have the same amount of parameters, the bias-only model has slightly higher inference cost than the adapter model whose adapter components are partially activated depending on the language.
But for simplicity, we will iterate based on the bias-only Conformer encoder model.

\subsection{Multilingual Decoder Architecture}

Besides using a single shared decoder, multi-head models \cite{pratap2020massively} that use different decoders for different languages/families can be used to add per task capacity. Similarly to \cite{pratap2020massively}, we use per language family decoders. Totally 5 families are used: Germanic, Italic, Arabic, Indo-Iranian and others. For comparison, we ensure the single decoder and the multi-decoder model have the same number of parameters: {\bf (1) single-decoder} has 6 layers of 768D LSTM with 3074 hidden units and {\bf (2) multi-decoder} has 5 decoders and each have 2 layers of 640D LSTM with 2048 hidden units. 
Both models have 354M parameters. At 200K steps, single-decoder yields of an average WER of 10.13\% compared to the multi-decoder's 10.28\%. This suggests given the same model size, it's more beneficial to use a single decoder which encourages more cross language/family sharing. 

To push the performance of our multilingual model, we further increase the 354M model to 500M by increasing the encoder's width from 512D to 640D and depth from 17 to 22 layers. It reaches an average WER of 9.63\% @200K and converges to 9.13\% @1.1M steps, outperforming the monolingual models. However, its training speed is less than 1/3 of the baseline 220M model due to the error back propagation through time for RNN models. This makes it unfavourable for further scaling up.

Transformer decoder \cite{Vaswani17} instead does not have the time recurrence constraint and has much high parallelism in training. With the same encoder architecture, we build a Transformer decoder model with totally 500M parameters, which leads to 12 Transformer layers with 768D model dim, 3072D hidden dim and 8 attention heads. It converges to a slightly higher WER of 9.26\% but with a training speed similar to the 220M baseline. We hence use Transformer decoders for following studies. 

\subsection{Scaling up with GShard}
\label{sec:1b_model}

\begin{table}[t]
\caption{1B-param model comparisons ({\bf L}: number of layers; {\bf W}: model dim; {\bf Loss}: training negative log perplexity, lower is better; {\bf Speed}: number of training examples per second). B0 is the baseline model and ``-'' indicates no change from B0.}
\hspace{-0.1in}
\begin{tabular}{lccccccc}
\toprule
\multirow{2}{*}{\tbh{Exp.}} & \multicolumn{2}{c}{\tbh{Encoder}} & \multicolumn{2}{c}{\tbh{Decoder}} & \multicolumn{2}{c}{\tbh{Training}} & \multirow{2}{*}{\tbh{WER$\downarrow$}} \\
\cmidrule(lr){2-3}\cmidrule(lr){4-5}\cmidrule(lr){6-7}
~ & \tbh{L} & \tbh{W} & \tbh{L} & \tbh{W} & \tbh{Loss$\downarrow$} & \tbh{Speed$\uparrow$} & {\scriptsize (@200K)} \\
\midrule
\midrule
B0 & 17 & 768 & 12 & 768 & 0.158 & 5530 & 10.36 \\
\midrule
E1 & 61 & - & - & - & 0.155 & 2352 & 10.13 \\
E2 & - & 1408 & - & - & 0.150 & \bf{3419} & 10.17\\
E3 & 33 & 1024 & - & - & 0.149 & 2975 & \bf{10.05} \\
E4 & 26 & 1152 & - & - & 0.151 & 3198 & 10.23\\
\midrule
E5 & - & - & 76 & - & \bf{0.143} & 2111 & 10.15\\
E6 & - & - & - & 1920 & 0.149 & 3170 & 10.48 \\
E7 & - & - & 22 & 1408 & 0.147 & 3204 & 10.37\\
\midrule
E8 & 22 & 1024 & 18 & 1152 & 0.147 & 2870 & 10.08\\
\bottomrule
\end{tabular}
\label{tbl:1bmodel}
\end{table}

In this experiment, we want to find the best way to further scale up the Conformer encoder and Transformer decoder model to 1B parameters. 
The set of experiments are listed in Table \ref{tbl:1bmodel}. Comparing E1 {\it vs.} E2 and E5 {\it vs.} E6, for both encoder and decoder, deeper model has better WER than wider models. However, deeper models are slow to train. Comparing E1-E4 with E5-E7, adding capacity to encoder does slightly better than decoder in terms of WER, however, large decoder tends to have better training loss. This might suggest the decoder's modeling task of the current speech training data is relatively simpler than the encoder's modeling task and larger decoder shows signs of over fitting. E4 that equally splits the additional capacity to width and depth \cite{kaplan2020scaling} does not work well on this task instead with more capacity allocated to depth (i.e. E3) we obtained the best WER. Lastly, E8, which first equally splits the capacity between encoder and decoder and then allocates more to depth, performs similarly to E3. 

E3 converges around 600K steps, roughly 10 epochs, halving the number of training iterations needed for the smaller models though each step runs longer. It is more data efficient. More importantly, it achieves an average WER of 9.07\%.  

\begin{figure*}[!t]
\hspace{-0.05in}
    \includegraphics[scale=0.39]{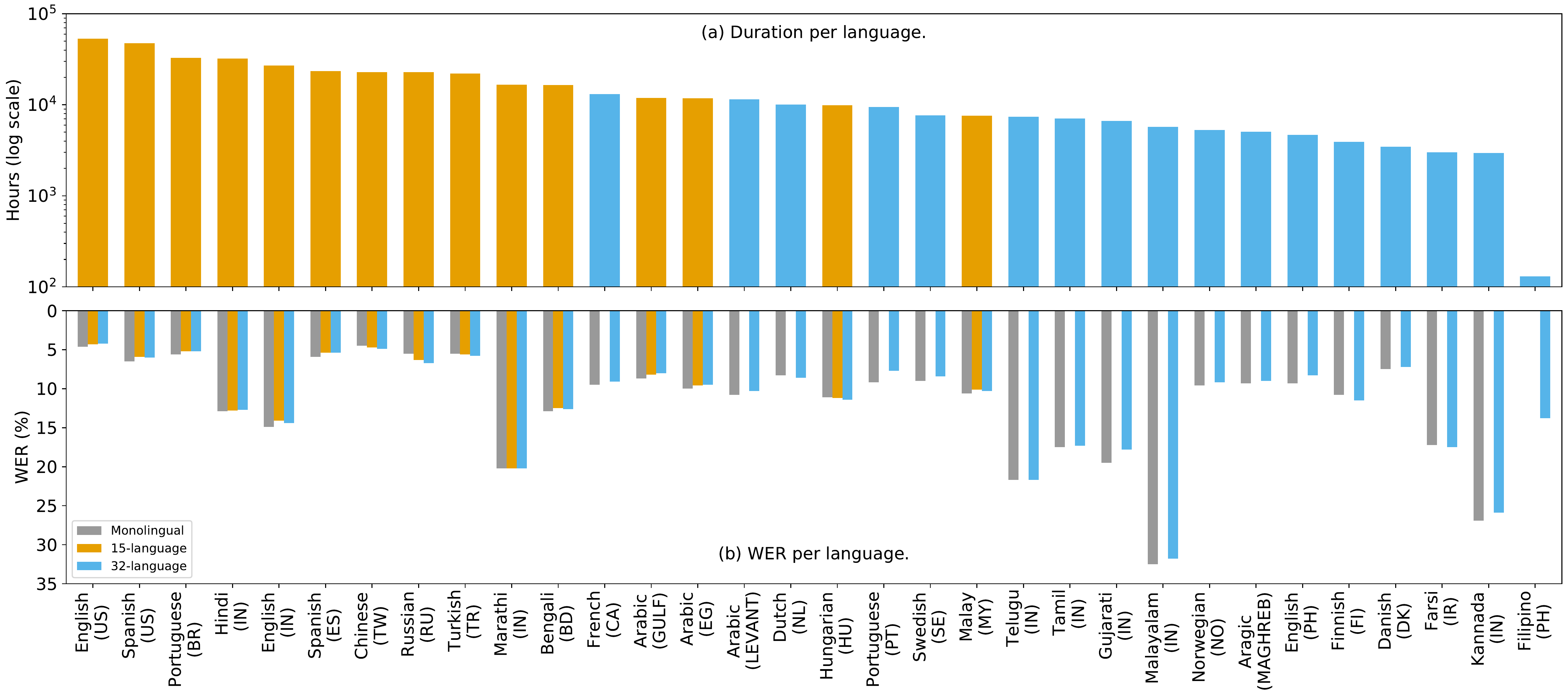}
    \caption{Extending the 15-language 1B model to 32 languages. (a) The data duration distribution per language. Yellow bars indicate the existing 15 languages and blue bars indicate the newly added languages. (b) WER (\%) comparisons between monolingual models (grey), the existing 15-language 1B model (yellow) and the same 1B model extended to 32 languages (blue).}
    \vspace{-0.15in}
    \label{fig:32lang}
\end{figure*}

\subsection{Towards 10B-Param Model}

Based on the previous experiments, we further scale up the model size towards 10B parameters by focusing more capacity on the encoder and depth. Specifically, we increased the E3 encoder depth from 33 to 86 and width from 1024 to 2048 and kept the decoder the same. It converges to 9.04\% at 330K steps ($\sim$6 epochs). Although the WER reduction comparing to the 1B model is marginal, the larger capacity leaves more room for scaling up from 15 languages to more in future.

Besides the performance gains, we find that large models tend to be more data and training cost efficient, {\it i.e.} they can reach the same level of performance with fewer optimization steps (Figure \ref{fig:final}(a)) and less TPU days (Figure \ref{fig:final}(b)). This is similar to the observation in \cite{kaplan2020scaling}.
We did not scale beyond 10B mainly due to the slow training speed with current hardware. As shown in Figure~\ref{fig:final}(a), 10B model has better sample efficiency than 1B, {\it i.e.} less training epochs are needed to reach the same WER, but the longer TPU time required per step makes it impractical for now. Sparse models \cite{lepikhin2020gshard} have been found to scale up more efficiently, which will be explored in future work.

\subsection{Human-in-the-loop Data Balancing}
\label{sec:balancing}

For simplicity, we only compared the average WER across models. To understand how they do on each language, we plot the breakdown in Figure \ref{fig:final}(c) for the monolingual, 220M, 500M, 1B and 10B at convergence. Larger models win over monolingual models on most languages; however, there are languages they lag behind, especially on Russian (RU). We suspect this is because of the unbalanced data distribution. The amount of data per language mainly depends on the data collection project and has less consideration of the language complexity itself. To validate this, we take the 1B model and increase the mixing ratio for Russian (RU) to 0.4 and all the others with the same weights to continue training for another 130K steps. This reduces the WER on Russian (RU) from 6.3\% to 5.1\% which wins over the monolingual's 5.5\%. As we still maintain a small weight for others, no clear degradation is observed and the average WER is further reduced from 9.07\% to 8.87\%.  This suggests it is beneficial to find a better data mixing ratio or a schedule of mixing ratios for multilingual models, which will be explored in future.

\begin{table*}[t]
\caption{Test set WER (\%) comparisons on the multilingual Librispeech dataset. Languages that are used in the 15-language 1B model training are grouped under ``\textbf{Seen Languages}'' and the remaining ones are in ``\textbf{Unseen Languages}''.}
\centering
\begin{tabular}{lccccccccc}
\toprule 
\multirow{2}{*}{\tbh{Exp.}} & \multicolumn{3}{c}{\tbh{Seen Languages}} & \multicolumn{5}{c}{\tbh{Unseen Languages}}  & \multirow{2}{*}{\tbh{Avg.}} \\
\cmidrule(lr){2-4}\cmidrule(lr){5-9}
~ & \tbh{en} & \tbh{es} & \tbh{pt} & \tbh{de} & \tbh{nl} & \tbh{fr} & \tbh{it} & \tbh{pl} & ~\\
\midrule
\midrule
Monolingual \cite{Pratap2020MLSAL} & 6.76 & 6.68 & 20.52 & 7.10 & 13.09 & 6.58 & 11.78 &  21.66 & 11.8 \\
\quad + 5-gram LM \cite{Pratap2020MLSAL} & 5.88 & 6.07 & 19.49 & 6.49 & 12.02 & 5.58 & 10.54 & 20.39 & 10.8 \\
\midrule
XLSR \cite{conneau2020unsupervised} & - & 6.3 & \bf{14.7} & 7.0 & 10.8 & 7.6 &  10.4 & 17.2 & 10.6 \\
\midrule
B0 (370M, random init.) & 6.1 & 5.8 & 16.2 & 5.5 & 11.9 & 6.9 &  11.9 &  15.4 & 10.0 \\
B0 (370M, 15-language model init.) & 6.6 & 4.7 & 15.5 & 5.0 & 11.1 & 6.1  & 10.1 &  10.9 & 8.8 \\
E3 (1B, 15-language model init.) & \bf{5.8} & \bf{4.2} & 15.2 & \bf{4.3} & \bf{9.9} & \bf{4.9}  & \bf{8.8} &  \bf{10.4} & \bf{7.9} \\
\bottomrule
\end{tabular}
\label{tbl:mls}
\end{table*}

\subsection{Extending to More Languages}

One of the challenges in developing multilingual systems is how to extend the models to new languages. In this work, we take a naive approach by continuing the model training on a combination of data from the existing languages and new languages. Specifically, we bring data from 17 additional languages and merge them with data from the existing 15 languages listed in Table~\ref{tbl:data}. Similarly, the data is mixed based on its natural distribution. This brings in additional 80M utterances, which leads to a total of 312.2M utterances and 466.4K hours of speech. Detailed per language duration information is depicted in Figure~\ref{fig:32lang} (a). The newly added languages are in the middle resource group, except for Filipino (PH) which has only 130.6 hours of training data. On the model input side, we expand the 1-hot language ID vector from 16D to 32D to cover the newly added languages. On the model output side, the new languages slightly increase the output grapheme vocabulary from 3328 to 3712. With these changes, we initialize from the 1B model (E3 Table~\ref{tbl:1bmodel}) trained on the existing 15 languages with an average WER of 9.07\% ({\it c.f.} Section~\ref{sec:1b_model}) and continued training on the combined 32 languages for another 600K steps. We didn't observe additional gains by training longer. The WER comparisons among the monolingual models (each has 142M model parameters), the 15-language 1B model and this 32-language 1B model are shown in Figure~\ref{fig:32lang} (b). On the 15-language set (Table~\ref{tbl:data}), the 32-language model has an average WER of 9.15\%, which is better than monolingual's 9.29\% and is close to the 15-language model's 9.07\%. On all the 32 languages, the single 1B model achieves an average WER of 11.57\% which is  better than the 11.87\% WER obtained by training 32 monolingual 142M-param models. Among those monolingual models, we failed to build one for Filipino (PH) due to over-fitting. But the 1B-param multilingual model works seamlessly well and achieves a WER of 13.8\%,  without up-sampling its 130.6 hours of training data which is only 0.03\% of the total training data. In summary, by continuously training the model with data from new languages, we can maintain similar performance on existing languages and obtain comparable performance on new ones. With even larger capacity and better data balancing techniques, it is potential to further improve the multilingual model performance, which will be explored in future. 

\subsection{Extending to Different Domains}

Besides language differences, domain mismatch is another challenging problem for practical speech recognition systems, such as voice action speech vs. read speech, short-form speech vs. long-form speech. In this section, we are interested in whether these large-scale multilingual models can boost performance for models on a different domain. We took our 15-language models to seed the training of a multilingual model on the Multilingual Librispeech dataset \cite{Pratap2020MLSAL}, which contains 8 languages, namely English ({\bf en}), German ({\bf de}), Dutch ({\bf nl}), French ({\bf fr}), Spanish ({\bf es}), Italian ({\bf it}), Portuguese ({\bf pt}) and Polish ({\bf pl}). Five of them are not present in our 15-language setup (Table~\ref{tbl:data}) and the data is read speech with longer duration of 10$\sim$20 seconds per utterance. In this experiment, we used a 2-layer 768D LSTM with 3072 hidden units based RNN-T \cite{graves2012sequence} decoder instead of the Transformer-based LAS decoder because attention models tend to have higher deletion errors on long-form data. We use a global batch size of 1024 with 128 TPUs on this dataset. We first took the same encoder as B0 (Table~\ref{tbl:1bmodel}) and train from random initialization which gives us an average WER of 10.0\%. We then initialize the encoder from the B0 model trained on the 15-language voice search data, which converges to an average WER of 8.8\%. This shows that positive knowledge transferring happened from an existing multilingual model trained on data from different languages and different domains to this new model via simple continuous training. Specifically, a 12\% relative WER reduction on the multilingual Librispeech task is obtained. From the per-language WER breakdowns in Table~\ref{tbl:mls}, the pre-trained multilingual model not only improves the performance on the languages seen during the pre-training (mainly domain mismatches); it also improves the quality on the 5 languages not seen previously (both language and domain mismatches). This suggests that the representations learned in the multilingual training generalize well to unseen languages and domains. English is the only language that the fine-tuned model does worse than the randomly initialized one. This could be a capacity problem as English is the largest language on both the 15-language setup and the multilingual Librispeech task and the fine-tuning model needs to capture the variations from both domains. We then used the existing 15-language 1B-param model (E3) to seed the training and we can further reduce the average WER to 7.9\%, another 10\% relative WER reduction. This also fixes the regression on English. To our knowledge, our results are the best reported numbers so far on this multilingual Librispeech dataset in the literature \cite{Pratap2020MLSAL, conneau2020unsupervised}.

%% file: 5_concl.tex
\section{Conclusions}
\label{sec:concl}

In this work, we investigate the problem of building multilingual end-to-end ASR models on high resource languages with large scale datasets, where language interference becomes more prominent. We address this problem by scaling up model capacities and empirically show that we can build models up to 10B parameters. With larger models, we have observed consistent performance gains. Moreover, the larger models are more sample and training cost efficient, {\it i.e.} requiring less training optimization steps and TPU time, though each training step of giant models runs longer. With increased capacities, we can build a single multilingual model that outperforms the monolingual models on high resource languages on a large scale multilingual dataset. We do see on some languages the multilingual model is still lagging behind. Empirical evidence suggests it is a data balancing problem, which will be investigated in future. Furthermore, preliminary results have shown that such large-scale multilingual models can be continuously trained to expand to new languages and new domains.